\documentclass{article}

\usepackage{microtype}
\usepackage{graphicx}
\usepackage{multirow}
\usepackage{subcaption}
\usepackage{booktabs} 
\usepackage{array}     
\usepackage{hyperref}

\usepackage[accepted]{icml2025}

\usepackage{amsmath}
\usepackage{amssymb}
\usepackage{mathtools}
\usepackage{amsthm}
\usepackage{makecell}
\usepackage[capitalize,noabbrev]{cleveref}

\usepackage{xcolor}
\usepackage{soul}
\sethlcolor{yellow}
\theoremstyle{plain}

\theoremstyle{definition}

\theoremstyle{remark}

\usepackage[textsize=tiny]{todonotes}

\icmltitlerunning{Adviser-Actor-Critic}

\begin{document}

\twocolumn[
\icmltitle{Adviser-Actor-Critic: Eliminating Steady-State Error in \\
            Reinforcement Learning Control}



\icmlsetsymbol{equal}{*}

\begin{icmlauthorlist}
\icmlauthor{Donghe Chen}{sch}
\icmlauthor{Yubin Peng}{sch}
\icmlauthor{Tengjie Zheng}{sch}
\icmlauthor{Han Wang}{sch}
\icmlauthor{Chaoran Qu}{sch}
\icmlauthor{Lin Cheng}{sch}
\end{icmlauthorlist}

\icmlaffiliation{sch}{School of Astronautics, Beihang University, 100191 Beijing, China}
\icmlcorrespondingauthor{Lin Cheng}{chenglin5580@buaa.edu.cn}

\icmlkeywords{Reinforcement Learning, Robotics, Robust Control}

\vskip 0.3in
]



\printAffiliationsAndNotice{}  

\begin{abstract}
    High-precision control tasks present substantial challenges for reinforcement learning (RL) algorithms, frequently resulting in suboptimal performance attributed to network approximation inaccuracies and inadequate sample quality.These issues are exacerbated when the task requires the agent to achieve a precise goal state, as is common in robotics and other real-world applications.We introduce Adviser-Actor-Critic (AAC), designed to address the precision control dilemma by combining the precision of feedback control theory with the adaptive learning capability of RL and featuring an Adviser that mentors the actor to refine control actions, thereby enhancing the precision of goal attainment.Finally, through benchmark tests, AAC outperformed standard RL algorithms in precision-critical, goal-conditioned tasks, demonstrating AAC's high precision, reliability, and robustness.Code are available at: https://anonymous.4open.science/r/Adviser-Actor-Critic-8AC5.
 
\end{abstract}

\section{Introduction}

Achieving robust, high-precision control in complex systems is a key goal in fields like machine engineering and aerospace \cite{zhang_brief_2023,cheng_neural-network-based_2024,qu_dynamic-matching_2024}. As advanced machinery becomes more prevalent, mastering the control of these systems poses significant challenges for engineers. These challenges mainly involve accurately modeling system dynamics and managing nonlinear behavior, which can cause steady-state errors in control systems. In this paper, we propose a framework that combines control theory with deep reinforcement learning (RL) to achieve high-precision and robust control \cite{andrychowicz_hindsight_2018}.

Dynamic modeling is crucial for understanding robot behavior and designing control strategies. However, real-world systems often display nonlinear behavior, making it difficult to create accurate models. Additionally, the high-dimensional state space of robots can lead to complex interactions between components, further complicating control \cite{busoniu_reinforcement_2018,zhao_towards_2020,zhao_sim--real_2020,cao2023learning}.

To highlight these challenges, we discuss the attributes and limitations of existing control algorithms. (1). Classical control algorithms, such as Proportional-Integral-Derivative (PID) \cite{li_pid_2006,borase_review_2021} controllers, are effective for linear, single-input-single-output (SISO) systems but struggle with the nonlinear, multi-input-multi-output (MIMO) nature of complex robots \cite{liu2019review}. (2). Modern control algorithms like Model Predictive Control (MPC) \cite{darby_mpc_2012} excel at handling nonlinearities but require precise models and significant computational resources. (3). Reinforcement learning (RL) algorithms, such as Soft Actor-Critic (SAC) \cite{haarnoja_soft_2018} Proximal Policy Optimization (PPO) \cite{schulman_proximal_2017} Deep Deterministic Policy Gradient(DDPG) \cite{lillicrap_continuous_2019}, can operate without detailed system dynamics, making them suitable for complex and nonlinear problems \cite{hutter2016anymal,tessler2019action}. At same time, RL algorithms face issues like sparse rewards and approximation errors \cite{pathak_curiosity-driven_2017,ramakrishnan_discovering_2018,fujimoto2018addressing}, leading to suboptimal performance.

While RL algorithms offer the most flexibility and adaptability, they also present unique challenges. Efforts are focused on improving RL's control precision, robustness, and training efficiency to fully leverage its potential \cite{kiran_deep_2021,zhuang_comprehensive_2020,mnih2015human, silver2014deterministic}.

To improve control precision in reinforcement learning (RL), researchers have developed two main methods: reward shaping \cite{burda_exploration_2018,hu_learning_2020} and integrating integral error into observations \cite{tracey_towards_2023}. Reward shaping involves adding extra rewards to reduce steady-state errors, but designing effective reward structures can be complex. Integrating the integral of error as an observation element, inspired by PID control, aims to compensate for past errors. However, this method increases the observation dimensions and may not always enhance performance.

We propose the Adviser-Actor-Critic (AAC) framework, a model-free approach that integrates PID control with reinforcement learning (RL), with the following contributions:
(1). AAC employs PID controllers to minimize steady-state errors in robotic control tasks and demonstrates adaptability across various systems.(2). Building on this integration, we implement an adviser mechanism grounded in PID control theory, which enhances system robustness and adaptability by providing structured guidance for the actor-critic architecture.(3). The AAC framework has been evaluated in three goal-oriented environments, where it exhibits advantages in robustness, broad applicability, and efficient transition from simulation to real-world settings.

\section{Preliminary}
In this section, we introduce the preliminaries, such as problem modeling, feedback control theory and reinforcement learning approaches.
\subsection{Problem Modeling}\label{section:problem}
Consider a nonlinear time-invariant system described by:
\begin{equation}
   \begin{cases}
      \dot{\boldsymbol{s}} = \boldsymbol{f}(\boldsymbol{s}, \boldsymbol{a}), \\
      \boldsymbol{g}_a = \boldsymbol{\phi}(\boldsymbol{s}),\\
      \boldsymbol{g}_d = \boldsymbol{\psi}(t)
      \end{cases}
\end{equation}
where: $\boldsymbol{s} \in \mathbb{R}^n$ represents the state vector of the system at time $t$, $\boldsymbol{a} \in \mathbb{R}^m$ is the control input applied to the system, $\boldsymbol{g}_a \in \mathbb{R}^p$ is the output of the system, $\boldsymbol{g}_d \in \mathbb{R}^p$ is the desired output of the system, $\boldsymbol{f}: \mathbb{R}^n \times \mathbb{R}^m \to \mathbb{R}^n$ is a continuous vector field representing the nonlinear dynamics of the system, $\boldsymbol{\phi}: \mathbb{R}^n \to \mathbb{R}^p$ is a continuous function mapping the state to the system output. $\boldsymbol{\psi}: \mathbb{R} \to \mathbb{R}^p$ is a continuous function mapping time to desired the system output.

Our objective is to formulate a control policy $\boldsymbol{a} = \boldsymbol{\kappa}(\boldsymbol{s}, \boldsymbol{g}_a, \boldsymbol{g}_d)$ aimed at maximizing the reward function given by Equation (\ref{eq:reward_function}).
\begin{equation}
   \label{eq:reward_function}
   R = \sum_{t=0}^{\infty} r(\boldsymbol{s}, \boldsymbol{g}_a, \boldsymbol{g}_d, \boldsymbol{a})
\end{equation}
Here, the function $r(\boldsymbol{s}, \boldsymbol{g}_a, \boldsymbol{g}_d, \boldsymbol{a})$ encompasses both the reward for reducing the discrepancy between the vectors $\boldsymbol{g}_a$ and $\boldsymbol{g}_d$, and the cost savings or benefits associated with the execution of actions $\boldsymbol{a}$.

The foregoing encapsulates a generic portrayal within the domain of control theory. However, within the specialized context of goal-conditioned reinforcement learning (RL) environments\cite{liu_goal-conditioned_2022}, data are typically structured in a dictionary format, encompassing three key components:

\begin{equation}
   \text{data\_dictionary} = 
   \begin{bmatrix}
      \text{'observation'} & : & \boldsymbol{s} \\
      \text{'desired\_goal'} & : & \boldsymbol{g}_d \\
      \text{'achieved\_goal'} & : & \boldsymbol{g}_a
   \end{bmatrix}
\end{equation}

Here, the \textit{desired goal} ($\boldsymbol{g}_d$) is the task the agent must accomplish, which can be provided by the environment or generated intrinsically. The \textit{achieved goal} ($\boldsymbol{g}_a$) is the goal reached by the agent at the current state. To better align with common RL paradigms, we reformulate the data dictionary into an extended observation vector, denoted as $\boldsymbol{s}_e$, as shown in Equation \ref{eq:extended_observation}:

\begin{equation}
   \label{eq:extended_observation}
   \boldsymbol{s}_e = [\boldsymbol{s}, \boldsymbol{g}_a, \boldsymbol{g}_a - \boldsymbol{g}_d] \overset{\boldsymbol{e}=\boldsymbol{g}_d - \boldsymbol{g}_a}{=\!=\!=\!=\!=\!=} [\boldsymbol{s}, \boldsymbol{g}_a, -\boldsymbol{e}]
\end{equation}

\subsection{Markov Decision Process}
An agent interacts with an environment, where the environment is fully observable and consists of a set of states $S$, actions $A$, initial state distribution $p(s_0)$, transition probabilities $p(s_{t+1}|s_t,a_t)$, reward function $r: S \times A \to \mathbb{R}$, and discount factor $\gamma \in [0, 1]$. These components form a Markov Decision Process (MDP) represented as $(S, A, p, r, \gamma)$. The agent's policy $\pi$ maps states to actions, $\pi : S \to A$.
At the start of each episode, the agent samples an initial state $s_0$ from $p(s_0)$. During each time step $t$, the agent selects an action $a_t$ following $\pi$ when in state $s_t$. After taking action $a_t$, the environment returns a reward $r_t = r(s_t, a_t)$ and the next state $s_{t+1}$ is sampled from $p(\cdot | s_t, a_t)$. Rewards may be discounted by $\gamma$ at each time step.
The agent's primary objective is to maximize its total discounted reward, known as the return $R_t = \sum_{i=t}^{\infty}\gamma^{i-t}r_i$, across all episodes. This corresponds to optimizing the expected return $\mathbb{E}_{s_0}[R_0|s_0]$.

\subsection{Proportional-Integral-Derivative (PID) Control}

Proportional-Integral-Derivative (PID) control \cite{li_pid_2006} is a cornerstone feedback mechanism widely employed in industrial control systems and automation. PID controllers operate by calculating an error value as the discrepancy between a desired setpoint and the actual system output. The objective of the controller is to minimize this error through adjustments to the control inputs.

The performance of a PID controller is finely tuned by adjusting three key parameters:\textbf{Proportional Gain ($K_p$)} determines the reaction's strength to the current error, balancing the speed of response; \textbf{Integral Gain ($K_i$)} addresses the accumulation of past errors to minimize steady-state errors over time; and \textbf{Derivative Gain ($K_d$)} predicts future trends based on the current rate of change, thereby dampening oscillations and enhancing stability.

Through the meticulous adjustment of $K_p$, $K_i$, and $K_d$, PID controllers achieve a harmonious balance between the speed of the system's response, its overall stability, and the precision with which it can maintain the desired output. This adaptability makes PID control an indispensable tool in a wide range of applications, from temperature regulation to motion control in robotics.

\section{Related Works}

To address the challenges of robotic control, researchers have developed several methodologies to enhance reinforcement learning (RL), improve control precision, and increase robustness. Three prominent techniques are Reward Shaping, Integrator Feedback, and Hindsight Experience Replay (HER).

\textbf{Reward Shaping}:
Reward Shaping modifies the reward function to help the agent learn faster and more effectively. It adds an extra reward to encourage minimizing the steady-state error, the difference between the desired and actual states. This additional feedback helps the agent perform better relative to the target state\cite{burda_exploration_2018,hu_learning_2020}.

\textbf{Integrator Feedback}:
Inspired by classical control theory, Integrator Feedback incorporates the integral of past error signals into the agent's observations. This reduces steady-state errors by considering both the current error and its integral.  However, it can introduce complexity and instability, leading to overshoot and oscillatory behavior. While it enhances control precision, it should be implemented carefully to avoid negative impacts on system stability \cite{tracey_towards_2023}.

\textbf{Hindsight Experience Replay (HER)}:
HER is an experience replay technique that reinterprets failed trajectories as successful ones by considering different, actually achieved goals. This method is particularly useful in goal-conditioned RL problems. By treating failures as learning opportunities, HER improves the efficiency and effectiveness of RL in complex and varied settings, especially for precise control and goal-oriented tasks \cite{andrychowicz_hindsight_2018,moro_goal-directed_2022}.

Both HER and Reward Shaping can be effectively integrated into the Adviser-Actor-Critic (AAC) framework, enhancing its performance. These strategies complement AAC by enriching the quality of experience. However, integral feedback does not serve a purposeful role within the AAC architecture. AAC offers a sophisticated alternative to integral feedback, especially for complex control tasks.

\section{Adviser-Actor-Critic}
This section starts with a motivating example that showcases how integrating traditional control theory with reinforcement learning (RL) enhances high-precision control. We then introduce the Adviser-Actor-Critic (AAC) framework, which combines both approaches to address complex control issues. Next, we analyze error dynamics to refine control strategies, and design advisers to guide the system toward its goals, improving performance and precision.
\subsection{A motivating example}
\begin{figure}[htbp]
   \centering
   \includegraphics[width=0.50\textwidth]{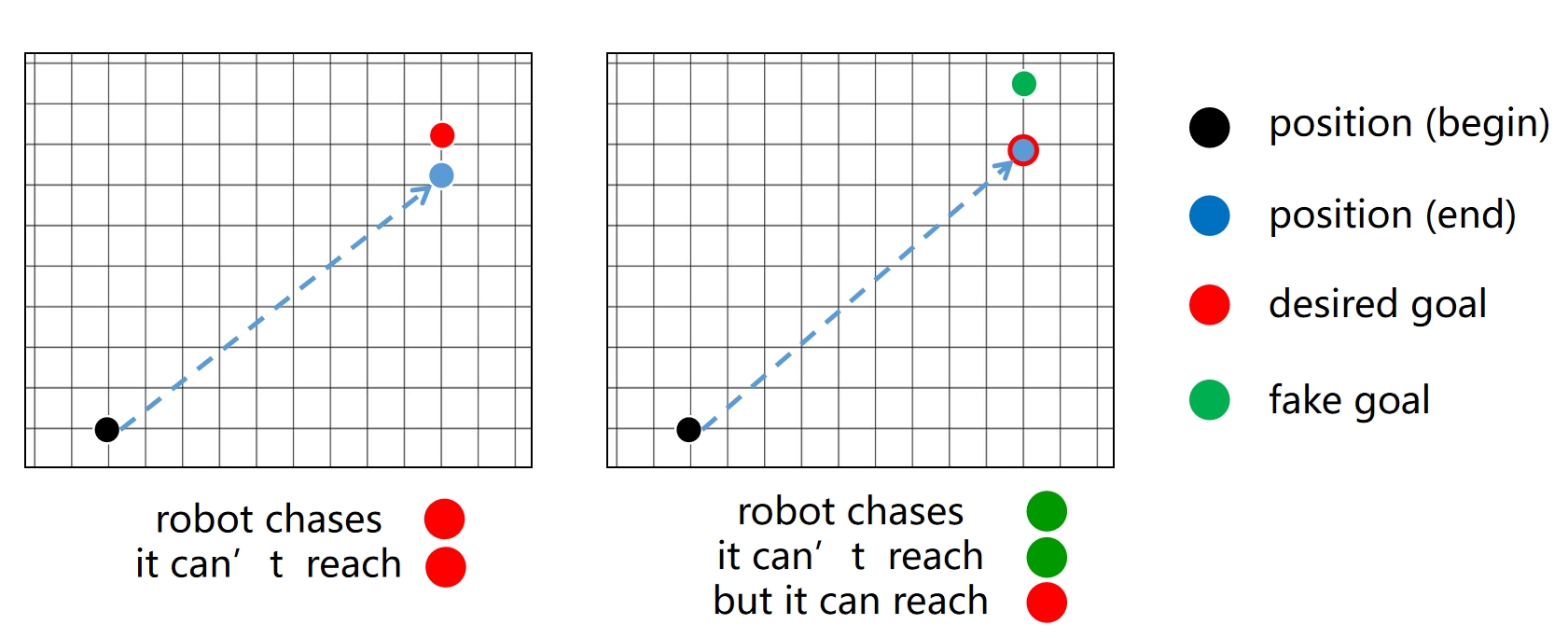}
   \caption{When a robot aims directly for a desired goal, control inaccuracies or model bias can prevent it from reaching the target precisely. However, by guiding the robot toward a strategically placed "fake goal," set by an adviser, it can effectively achieve the desired position. }
   \label{fig:basic_idea}
   \vskip -0.1in
\end{figure}

To grasp the fundamental concept of our approach, consider the scenario illustrated in Figure~\ref{fig:basic_idea}. Imagine a robot tasked with reaching a specific goal. Due to limitations in control precision or model bias, the robot cannot directly and precisely reach the target position. However, if the robot is guided by an advisor to pursue a "fake goal," it can employ this strategy to achieve the desired goal position, despite its inability to reach the fake goal itself. This concept indicates that even with control inaccuracies, the robot can attain precise control of the desired goal if it has an advisor capable of setting a suitable "fake goal."

Figure~\ref{fig:basic_idea} illustrates how traditional control methods, effective in stable environments, struggle with nonlinear and uncertain systems. Meanwhile, Reinforcement Learning (RL) adapts well to changes but can suffer from slow learning and suboptimal policies.The Adviser-Actor-Critic (AAC) framework addresses these issues by integrating the precision of control theory with RL's adaptability. It initially trains using deep RL, then refines performance through manually designed advisers, enhancing control precision without needing detailed system models or complex controllers.

\subsection{Algorithmic Framework}

\begin{figure*}[htbp]
   \centering
   \includegraphics[width=0.7\textwidth]{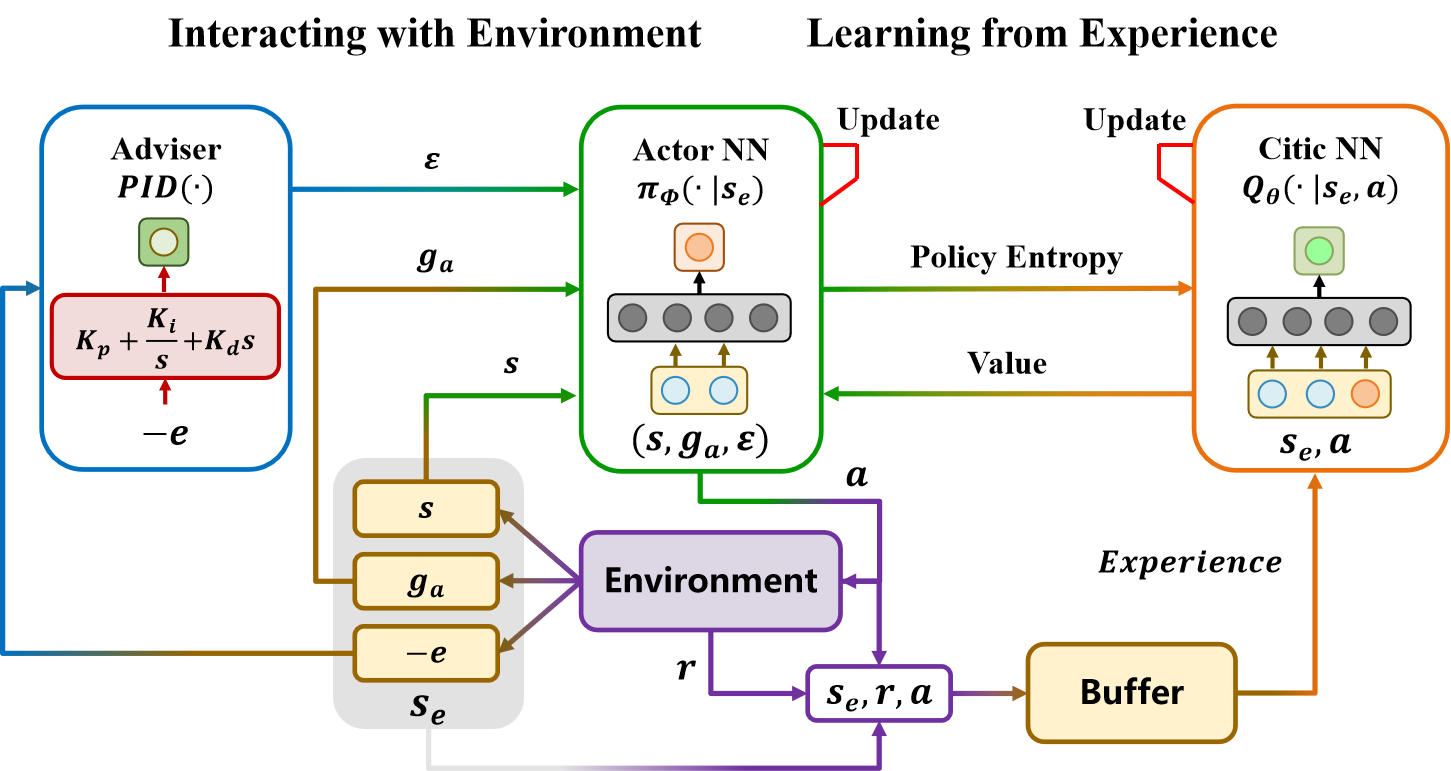}
   \caption{\textbf{Interact With Environment:} The Adviser outputs a synthetic error $\varepsilon$ to guide the Actor Neural Network's decision-making, alongside the achieved goal $g_a$ and current observation $s$. This framework integrates deep learning with classical control theory, enabling adaptive policy optimization and enhancing performance in dynamic environments. \textbf{Learn from Experience:} The Critic Neural Network estimates the state-action value function $Q_{\theta}(s_e, a)$, while the Actor Neural Network generates the policy $\pi_{\phi}(\cdot|s_e)$ based on the augmented observation $s_e$. Experiences are stored as tuples $(s_e, a, r, s_e')$ in an Experience Buffer to facilitate continuous learning. Feedback from the environment includes rewards and new observations.}
   \label{fig:framework}
   \vskip -0.1in
 \end{figure*}

The Adviser-Actor-Critic (AAC) framework, as illustrated in Figure~\ref{fig:framework}, offers a sophisticated approach to achieving high-precision control in environments characterized by dynamism and nonlinearity. It innovatively merges traditional control theory with reinforcement learning (RL), providing a balanced method that optimizes control precision and simplifies system design.

A distinctive aspect of the AAC framework is its implementation of a "fake goal" strategy through the Adviser. This mechanism guides the Actor towards more informed decision-making, addressing potential limitations of individual components operating in isolation. As shown in Figure~\ref{fig:framework}, the Adviser not only adjusts the target goal but also introduces a fake error signal $\varepsilon$, which aids the Actor Neural Network in refining its policy to maximize long-term rewards.

By integrating reinforcement learning with classical control theory, the AAC framework sets a new standard for intelligent control systems. Its adaptability and flexibility make it suitable for diverse applications, such as robotics, autonomous driving, and industrial automation, where maintaining precision and reliability under varying conditions is critical. The AAC framework thus represents a significant advancement in control system design, capable of delivering enhanced performance and robustness in complex, real-world scenarios.

\subsection{Adviser Design Based on PID Control}

In the previous section, we introduced the fundamental principles of the Adviser-Actor-Critic (AAC) framework and its approach to integrating traditional control theory with reinforcement learning for high-precision control. This section delves into the specific methods used to design advisers based on Proportional-Integral-Derivative (PID) control. PID control, widely used in industrial systems due to their robustness and adaptability, effectively minimize steady-state errors and improve system response by adjusting proportional, integral, and derivative gains. This flexibility makes them suitable for both linear and nonlinear systems, justifying their use in the AAC framework to enhance control precision and reliability.

\begin{figure*}[htbp]
   \centering
   \includegraphics[width=0.9\textwidth]{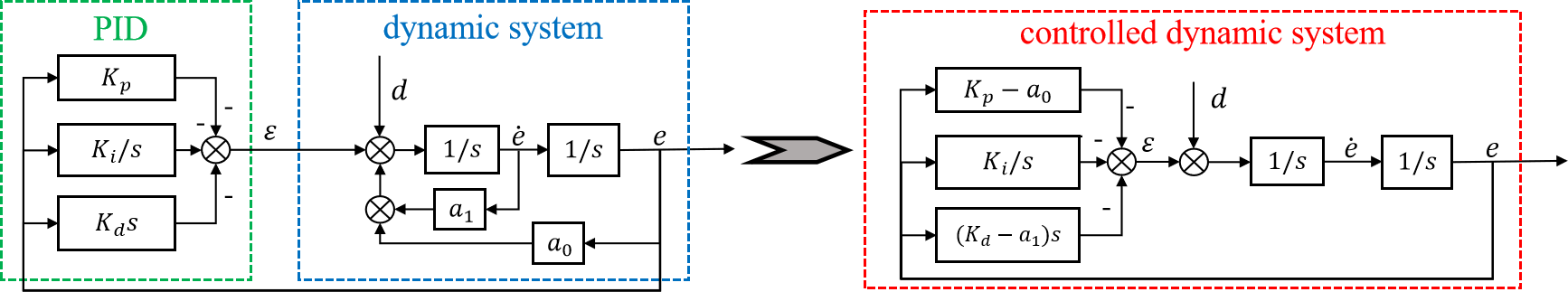}
   \caption{Within the dynamic system, PID controllers transform the actual error $ e $ into a fake error $ \varepsilon $. For simplified stability analysis, the PID controller parameters and dynamic system parameters are combined, adjusting $ K_p' = K_p - a_0 $ and $ K_d' = K_d - a_1 $. $ \varepsilon_i $ represents the adjusted error, while $ d_i $ denotes disturbances in the $ i $-th dimension, accounting for unmodeled dynamics. Notably, when $ K_p = 1 $, $ K_i = 0 $, and $ K_d = 0 $, the PID controller exhibits no additional effect, behaving as if no PID controller were present.}
   \label{fig:error_dynamic}
   \vskip -0.1in
\end{figure*}

Figure~\ref{fig:error_dynamic} illustrates the modeling techniques for each dimension of the system's dynamics, assuming the components are sufficiently decoupled (see Appendix~\ref{appendix:decoupled}). These models, aligned with traditional control theory, provide computationally efficient representations that approximate a wide range of physical behaviors, including oscillations and damping effects. 

A PID controller calculates the control from three key components: the current error $ e(t) $, the accumulated past errors (integral term), and the rate of change of the error (derivative term). The expression for a PID controller is given by:
\begin{equation}\label{eq:pid_controller}
   \epsilon(t) = -K_p e(t) - K_i \int_{0}^{t}e(\tau)\,d\tau - K_d \frac{d}{dt} e(t)
\end{equation}
where $ K_p $, $ K_i $, and $ K_d $ denote the proportional, integral, and derivative gains, respectively.

PID controllers are widely used in industrial control systems due to their robustness and adaptability. They offer a simple yet powerful method to minimize steady-state errors and improve system response. Their ability to handle both linear and nonlinear systems makes them highly versatile. By adjusting the proportional, integral, and derivative gains, PID controllers can be fine-tuned to achieve optimal performance across various applications. This flexibility and effectiveness justify their use in the AAC framework, where they enhance the precision and reliability of control actions.

\subsection{Adviser Performance Analysis}

To ensure the stability and reliability of the Adviser within the Adviser-Actor-Critic (AAC) framework, this section focuses on the stability analysis of the PID-controlled system. Stability is crucial for maintaining predictable system behavior under varying conditions. We use the Routh-Hurwitz stability criterion to determine the necessary conditions for system stability, ensuring that the Adviser remains stable under PID control. PID control is one effective method within a broader AAC strategy to enhance precision and reliability.

Figure~\ref{fig:error_dynamic} illustrates modeling techniques for each dimension of the system's dynamics, assuming decoupled components (see Appendix~\ref{appendix:decoupled}). This aligns with traditional control theory, using second-order models that provide computationally efficient representations suitable for classical control strategies like PID controllers.

\begin{equation}
   \label{eq:linear_system}
   \boldsymbol{\dot{x}} = \begin{bmatrix}
      1 & 0 \\
      a_0 & a_1 
      \end{bmatrix}\boldsymbol{x}
      +\begin{bmatrix}
         0 \\
         1 
         \end{bmatrix}\varepsilon_i
         +\begin{bmatrix}
            0 \\
            1
            \end{bmatrix}d_{i}
\end{equation}
where $\boldsymbol{x}$ denotes the state vector, which captures the system's state as $[e_i, \dot{e}_i]^T$, where $e_i$ is the error in the $i$-th dimension and $\dot{e}_i$ is its time derivative. The time derivative of the state vector, $\boldsymbol{\dot{x}}$, is represented by $[\dot{e}_i, \ddot{e}_i]^T$. Additionally, $\varepsilon_i$ signifies the simulated error introduced by the Adviser for the $i$-th dimension, while $d_i$ accounts for an external disturbance affecting the same dimension. The unknown dynamics of the system are encapsulated within $d_i$. This modeling approach is sufficiently general to encompass a wide range of system behaviors, making it suitable for diverse control applications.

\begin{figure}[htbp]
   \centering
   \includegraphics[width=0.45\textwidth]{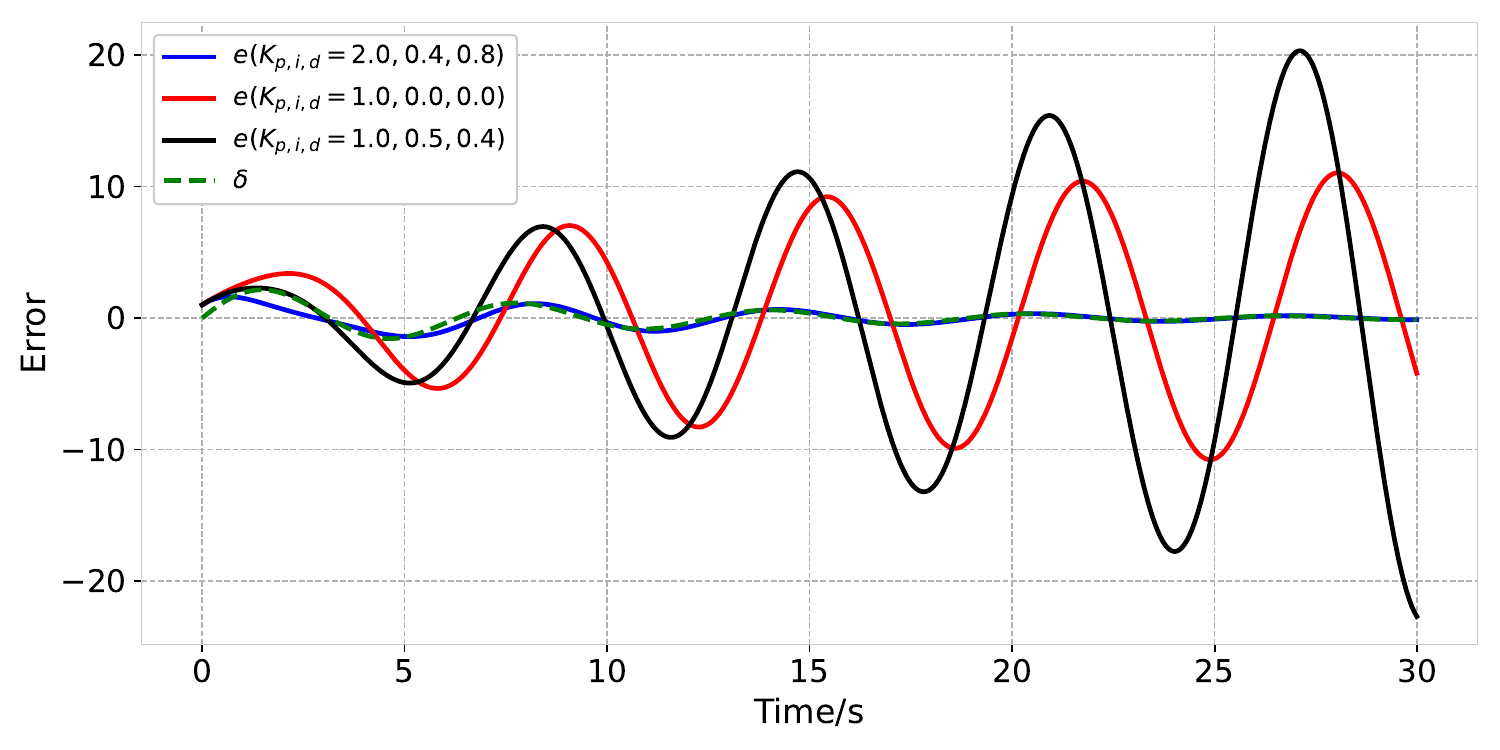}
   \caption{Response of a Second-Order System to Command under Different PID Parameters: Analyzing Asymptotic, Marginal, and Instability Conditions}
   \label{fig:pid_response}
   \vskip -0.1in
\end{figure}

Transitioning to the analysis of stability, it is essential to evaluate whether all poles of the closed-loop transfer function reside in the left half of the complex plane. As illustrated in Figure~\ref{fig:pid_response}, the closed-loop transfer function for the system under consideration can be expressed as:
\begin{equation}\label{eq:closed_loop_transfer_function}
\frac{E(s)}{D(s)} = \frac{1}{s^3 + K_d' s^2 + K_p' s + K_i}
\end{equation}
The characteristic equation, derived from the denominator of this transfer function, is given by $s^3 + K_d' s^2 + K_p' s + K_i = 0$. To assess the stability of the linear system, we employ the Routh-Hurwitz stability criterion, a widely recognized method that determines stability through the construction of the Routh array. For the specified characteristic equation, the Routh array is constructed as follows:
\begin{equation}\label{eq:routh_array}
\begin{array}{c|ccc}
s^3 & 1 & K_p' \\
s^2 & K_d' & K_i \\
s^1 & \frac{K_p' K_d' - K_i}{K_d'} & 0 \\
s^0 & K_i & 0  \\
\end{array}
\end{equation}

According to the Routh-Hurwitz stability criterion, the system is stable if and only if all elements in the first column of the Routh array are positive. Specifically, the stability conditions can be simplified into the following equations:
\begin{equation}
   \label{eq:stability_conditions}
   K_d' > 0 \quad K_i > 0 \quad  K_p' K_d' > K_i
\end{equation}

When Equation.~\ref{eq:stability_conditions} holds, all poles of the system will have negative real parts, signifying asymptotic stability. Conversely, failure to meet these conditions may result in instability or marginal stability.

Building on this foundation, we delve into the design principles of PID (Proportional-Integral-Derivative) controllers aimed at stabilizing dynamic systems. Utilizing the Routh-Hurwitz stability criterion as a key analytical tool, we methodically establish the necessary conditions for system stability. This framework is crucial for identifying appropriate PID parameters that ensure both optimal performance and robust stability margins.

\section{Experiments}
This section is organized as follows. First, we introduce multi-goal RL environments we use for the Experiments as our trainning procedure. Second, we compare the performance of SAC/SAC-HER with and without adviser. Third, we show how advisers leader actor to eliminate steady-state errors. Fourth, we compare different parameter sets of adviser. Finally, we show the results of the Experiments on physical robot.

\subsection{Environments}
Referring to gym goal-conditioned environment,we decided to use manipulation environments based on an existing hardware robot to ensure that the challenges we face corresponds as closely as possible to the real world. We use mass-spring-damper system, robotics arm and quadcopter in next three Experiments.

\begin{figure}[htbp]
   \centering
   \begin{subfigure}[b]{0.15\textwidth}
       \includegraphics[width=\textwidth]{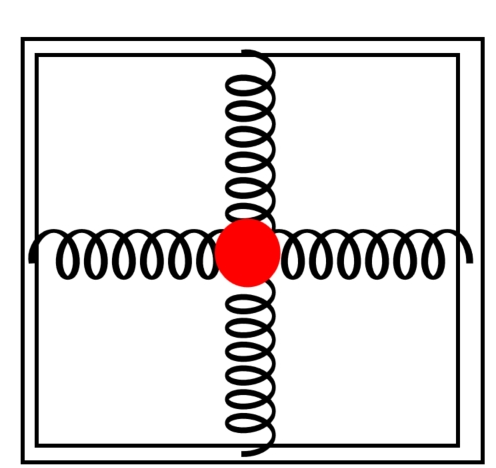}
   \end{subfigure}
   \begin{subfigure}[b]{0.15\textwidth}
       \includegraphics[width=\textwidth]{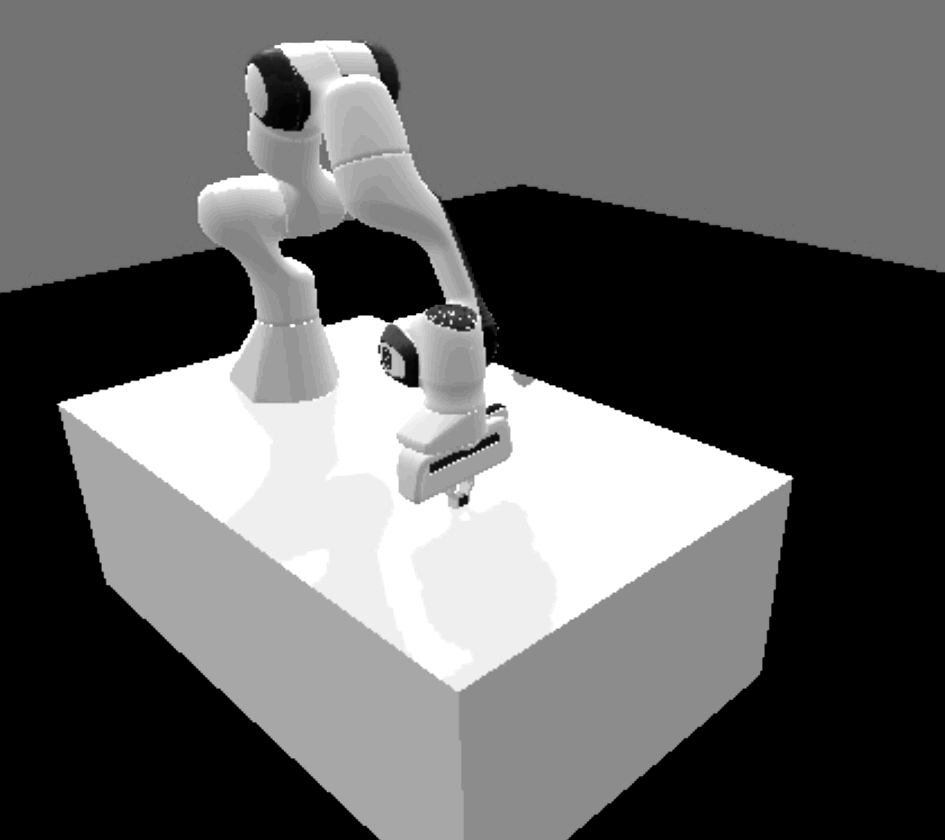}
   \end{subfigure}
   \begin{subfigure}[b]{0.15\textwidth}
       \includegraphics[width=\textwidth]{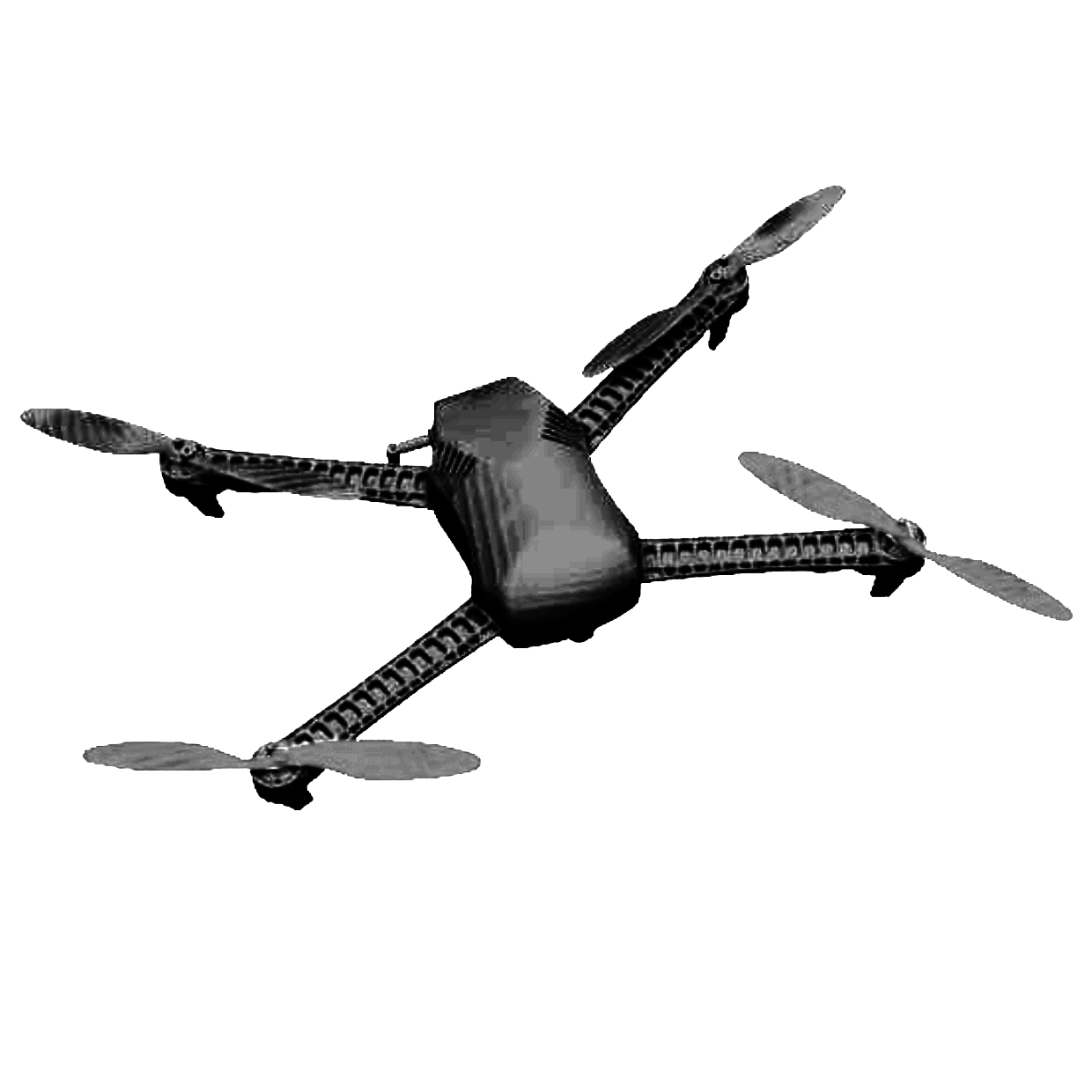}
   \end{subfigure}
   \caption{Different environments: Position Control of a Mass-Spring-Damper System (left), Robotics Arm Fetch (middle) and Velocity Control of a Quadcopter (right).}
   \label{fig:three graphs}
   \vskip -0.1in
\end{figure}

We evaluate the AAC on three tasks, as detailed below. It is represented as Multi-Layer Perceptrons (MLPs) with Sigmoid Linear Unit (SiLU) activation functions, and training is performed using the Soft Actor-Critic (SAC) algorithm with the Adam optimizer. Both the environment settings and the algorithm configurations are detailed in Appendix~\ref{appendix:experiment}.

\begin{itemize}
   \item \textbf{Position Control of a Mass-Spring-Damper System:} The goal is to apply an external force to move and maintain a mass at a desired position, adjusting based on current position and velocity.
   \item \textbf{Robotics Arm Fetch:} A robotic arm performs complex spatial movements and object manipulation using multiple independently movable joints and an end effector (e.g., gripper).
   \item \textbf{Velocity Control of a Quadcopter:} The quadcopter aims to fly at predetermined velocities and directions in 3D space, controlling pitch, roll, yaw, and vertical speed.
\end{itemize}

\subsection{How Does the Adviser Eliminate Steady-State Errors?}

\begin{figure}[htbp]
\centering
\includegraphics[width=0.48\textwidth]{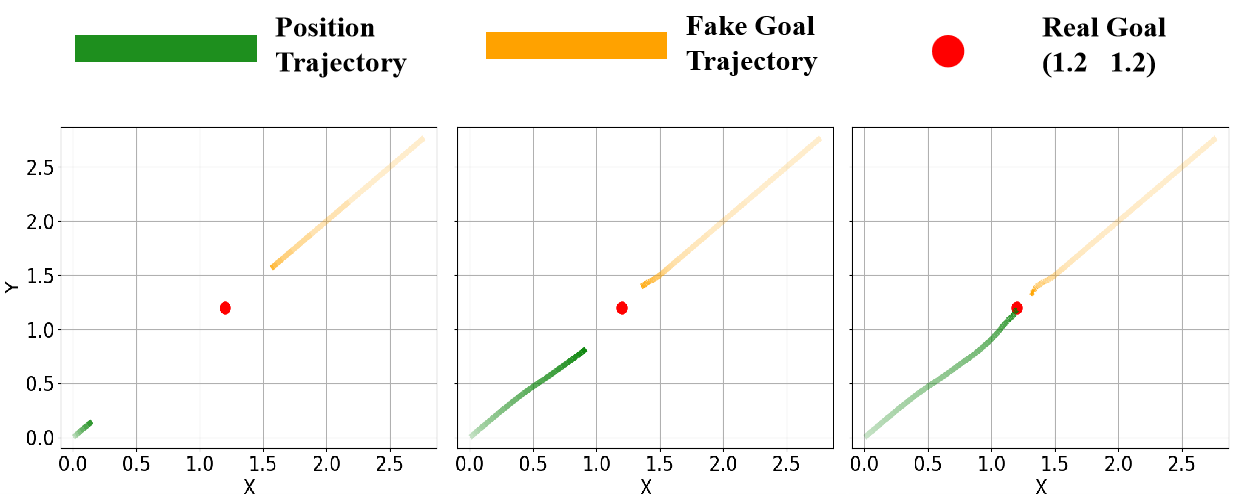}
\caption{Illustration of how the adviser uses intermediate goals to guide the robot towards the desired target. Starting from position (0, 0), the robot aims for a point at (1.2, 1.2). Although the intermediary targets do not perfectly align with the final destination, they serve as strategic waypoints that enable continuous improvement in the robot's trajectory, effectively minimizing steady-state errors over time.}
\label{fig:eliminateError}
\vskip -0.1in
\end{figure}

To comprehend how the adviser mitigates steady-state errors, one can examine Figure~\ref{fig:eliminateError}, which visually demonstrates the process. The adviser introduces strategically placed intermediate goals, or "fake" goals, to direct the robot toward the ultimate desired goal. Even though these intermediary targets may not coincide exactly with the final destination, they play a crucial role in ensuring the robot constantly refines its path and actions.

By setting up these waypoints, the adviser prompts the robot to make incremental adjustments to its course, which helps in overcoming any systematic deviations that could lead to steady-state errors. This method ensures that despite any initial discrepancies between the fake and desired goals, the robot can progressively converge on the correct path. As a result, the robot's navigation becomes more precise, and it can achieve its intended destination with significantly reduced persistent errors. 

\subsection{Does the Adviser Improve Performance?}

\begin{figure*}[htbp]
   \centering
   \includegraphics[width=\textwidth]{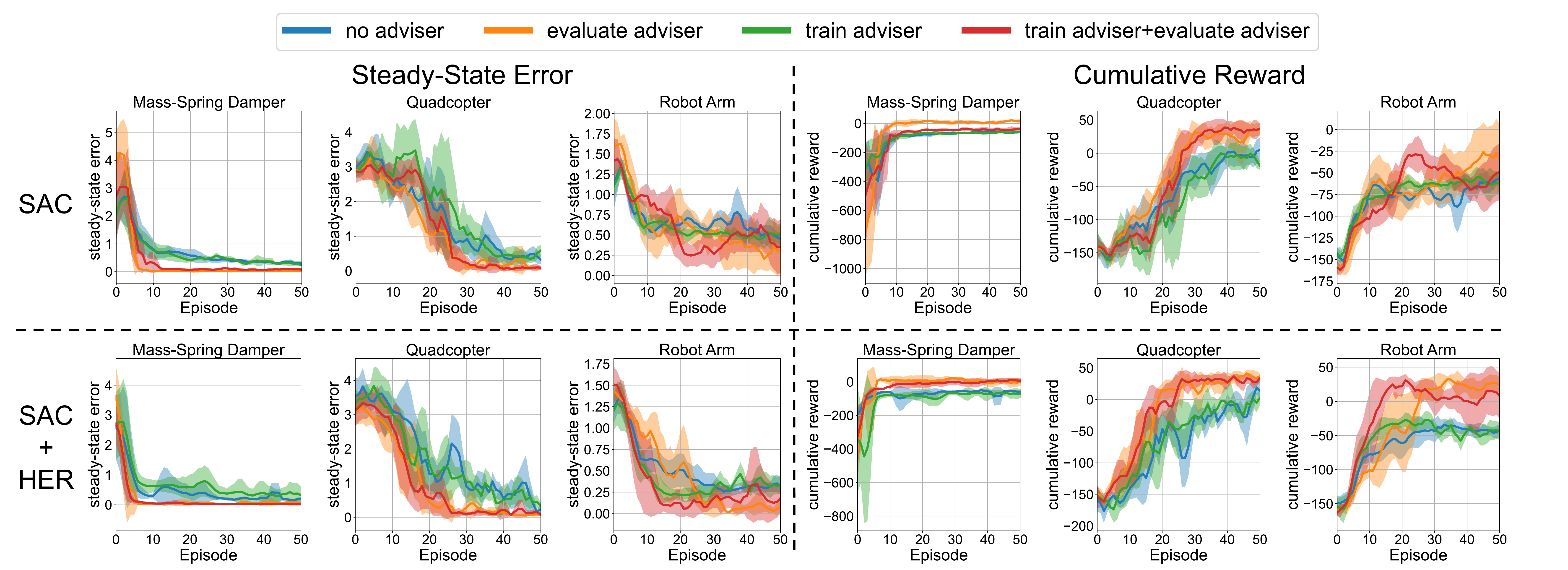}
   \caption{Performance comparison of four adviser configurations under SAC and SAC+HER across three environments. The adviser reduces steady-state errors and increases cumulative rewards, demonstrating enhanced stability and accelerated training.}
   \label{fig:combine}
   \vskip -0.1in
\end{figure*}

To determine whether the inclusion of an adviser can enhance performance, we conducted a detailed evaluation using two variants of the Soft Actor-Critic (SAC) algorithm: standard SAC and SAC with Hindsight Experience Replay (SAC+HER). These evaluations were carried out both with and without adviser support across all three experimental environments, allowing us to systematically analyze the adviser's impact on learning dynamics and overall performance.

Figure~\ref{fig:combine} provides a comprehensive overview of how adviser support significantly boosts the performance of both SAC and SAC+HER algorithms across various dimensions:

\begin{enumerate}
   \item \textbf{Enhanced Stability and Precision}: As illustrated in the left panel of Figure~\ref{fig:combine}, the presence of an adviser notably reduces steady-state errors for both SAC and SAC+HER. Notably, SAC+HER with adviser guidance achieves markedly lower errors compared to SAC without an adviser, underscoring the synergistic benefits of combining HER with advisory mechanisms.
   
   \item \textbf{Increased Cumulative Rewards and Learning Efficiency}: The right panel of Figure~\ref{fig:combine} highlights that adviser support leads to higher cumulative rewards for both algorithms and accelerates convergence to optimal policies. Across all three environments, SAC+HER with adviser support consistently outperforms other configurations, delivering superior rewards and more efficient learning processes.
\end{enumerate}

In summary, the integration of an adviser into the SAC and SAC+HER frameworks not only improves stability and precision but also enhances cumulative rewards and expedites the learning process, as evidenced by our experimental results. This suggests that advisers play a pivotal role in optimizing the performance of reinforcement learning algorithms in complex and dynamic environments.

\subsection{Deployment on a Real-World Quadcopter}

To thoroughly evaluate the effectiveness of our SAC (Soft Actor-Critic)-based attitude control algorithm, we conducted real-world experiments deploying the trained models onto an actual quadcopter system. The primary objective was to compare the performance and stability of the quadcopter when using the SAC algorithm both with and without advisory guidance during the training phase. The quadcopter's ability to maintain stable flight and respond effectively to external disturbances or deliberate control inputs is critical for ensuring safe and reliable operation in practical scenarios.

\begin{figure}[htbp]
   \centering
   \begin{subfigure}{0.13\textwidth}
       \includegraphics[width=\textwidth]{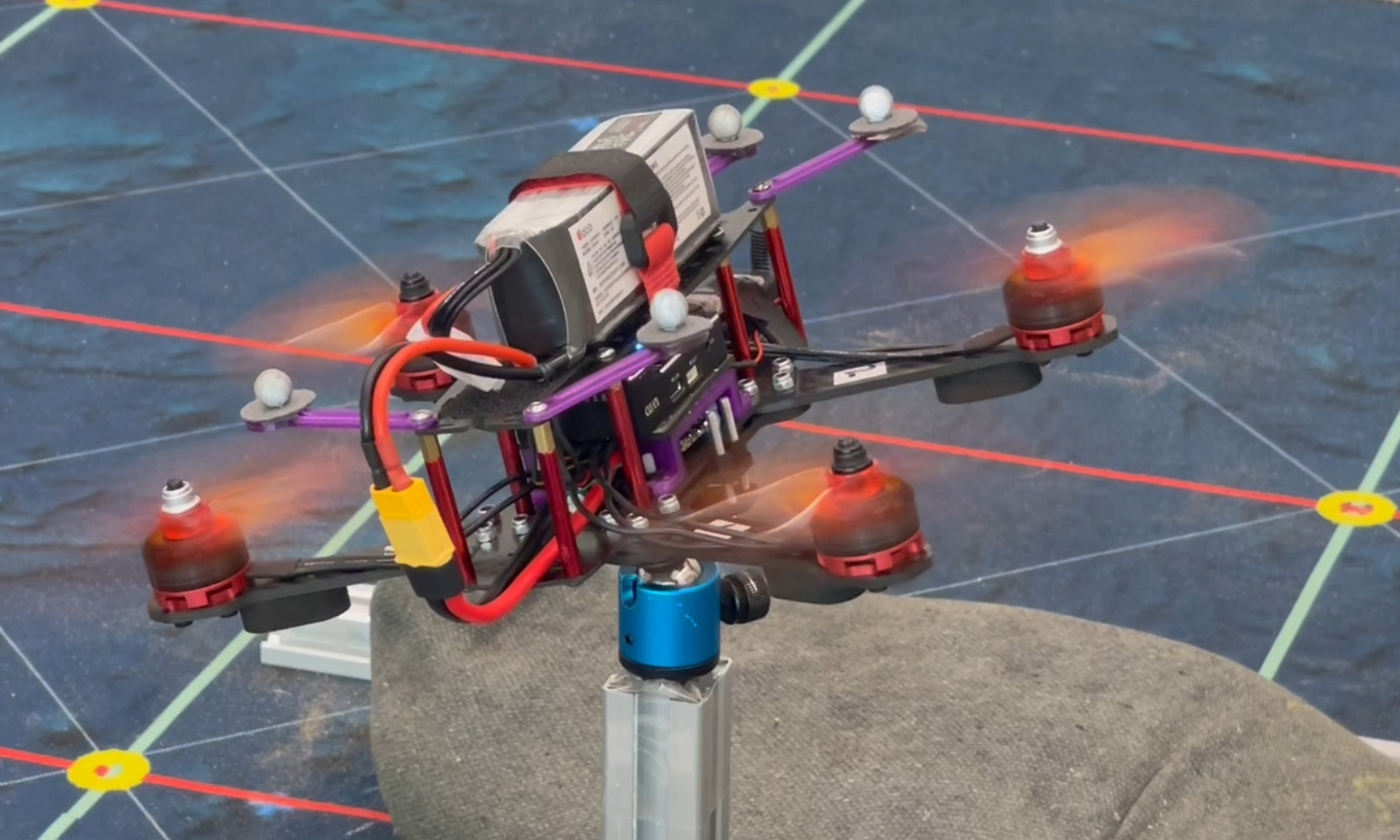}
       \caption{Tilted Left}
       \label{fig:quadcopter_left}
   \end{subfigure}
   \hspace{0.02\textwidth}
   \begin{subfigure}{0.13\textwidth}
       \includegraphics[width=\textwidth]{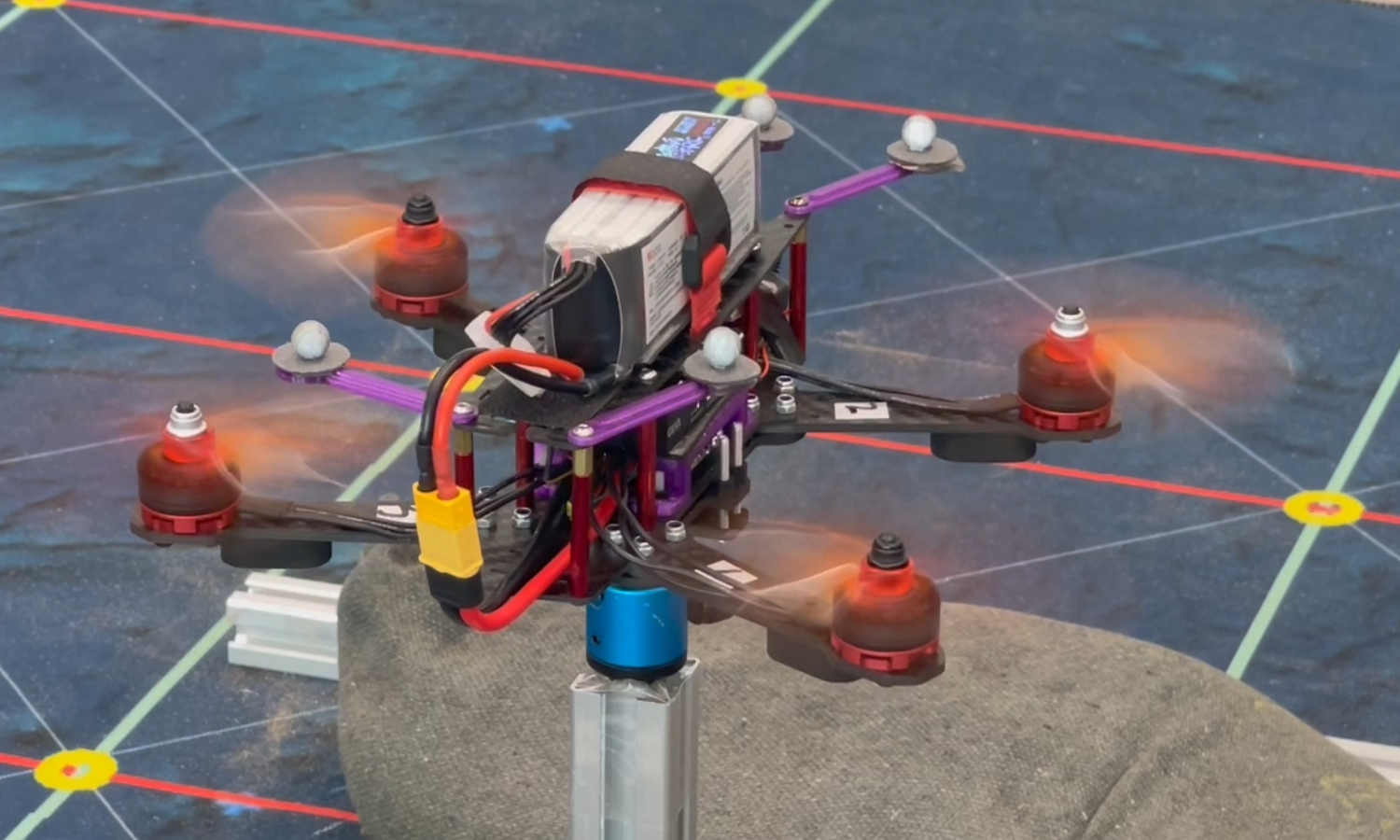}
       \caption{Neutral}
       \label{fig:quadcopter_neutral}
   \end{subfigure}
   \hspace{0.02\textwidth}
   \begin{subfigure}{0.13\textwidth}
       \includegraphics[width=\textwidth]{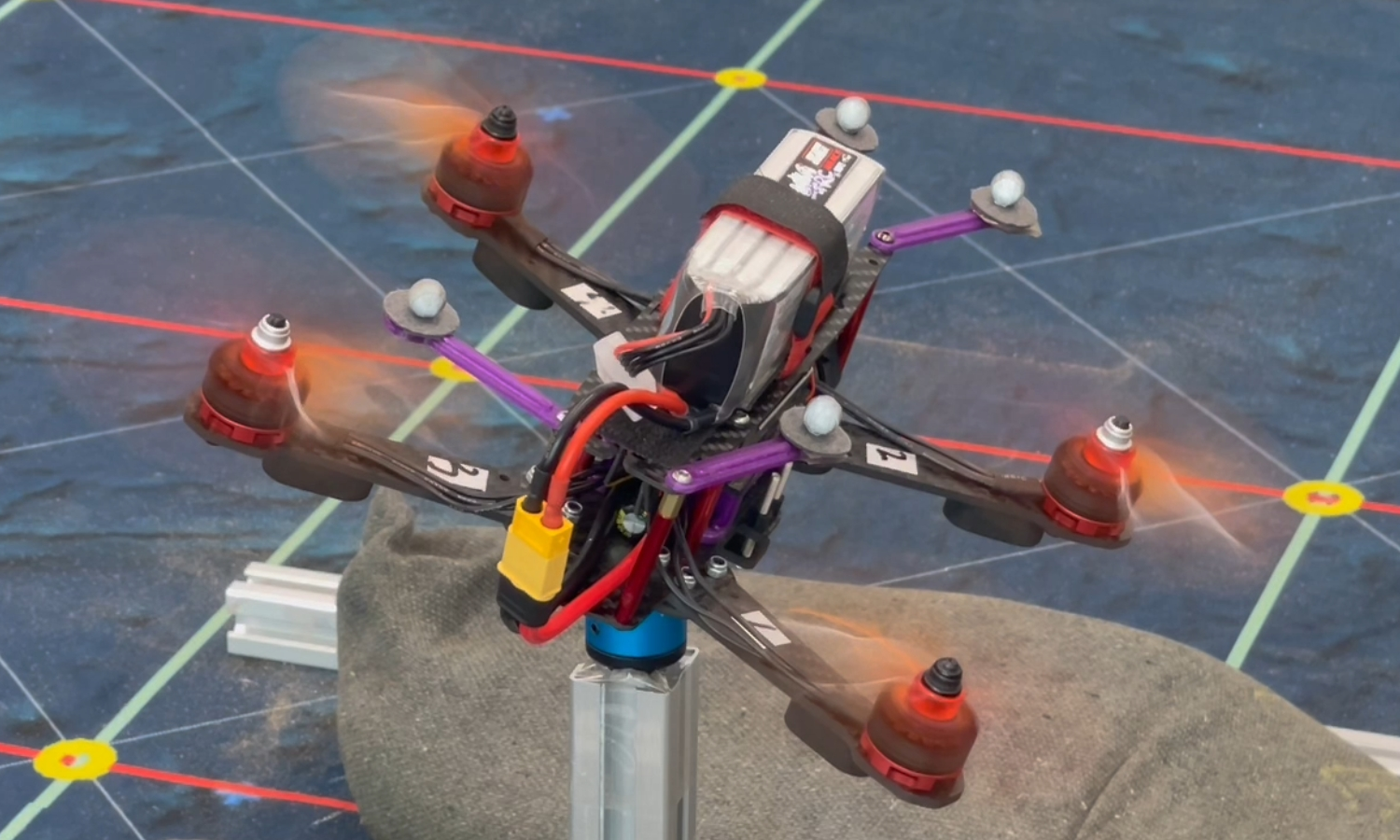}
       \caption{Tilted Right}
       \label{fig:quadcopter_right}
   \end{subfigure}
   \caption{Illustration of the quadcopter's orientations. The quadcopter, with a takeoff weight of 1.40 kg and an axle distance of 35.0 cm, uses a Pixhawk4 for low-level control and ROS for high-level commands. Indoor flights utilize a motion capture system for precise positioning.}
   \label{fig:quadcopter_poses}
   \vskip -0.1in
\end{figure}

The visual representation of the quadcopter's postures under different conditions can be seen in Figure~\ref{fig:quadcopter_poses}. This figure captures the vehicle at various angles, showcasing its dynamic range of motion.

\begin{figure}[htbp]
   \centering
   \begin{subfigure}[b]{0.35\textwidth}
       \includegraphics[width=\textwidth]{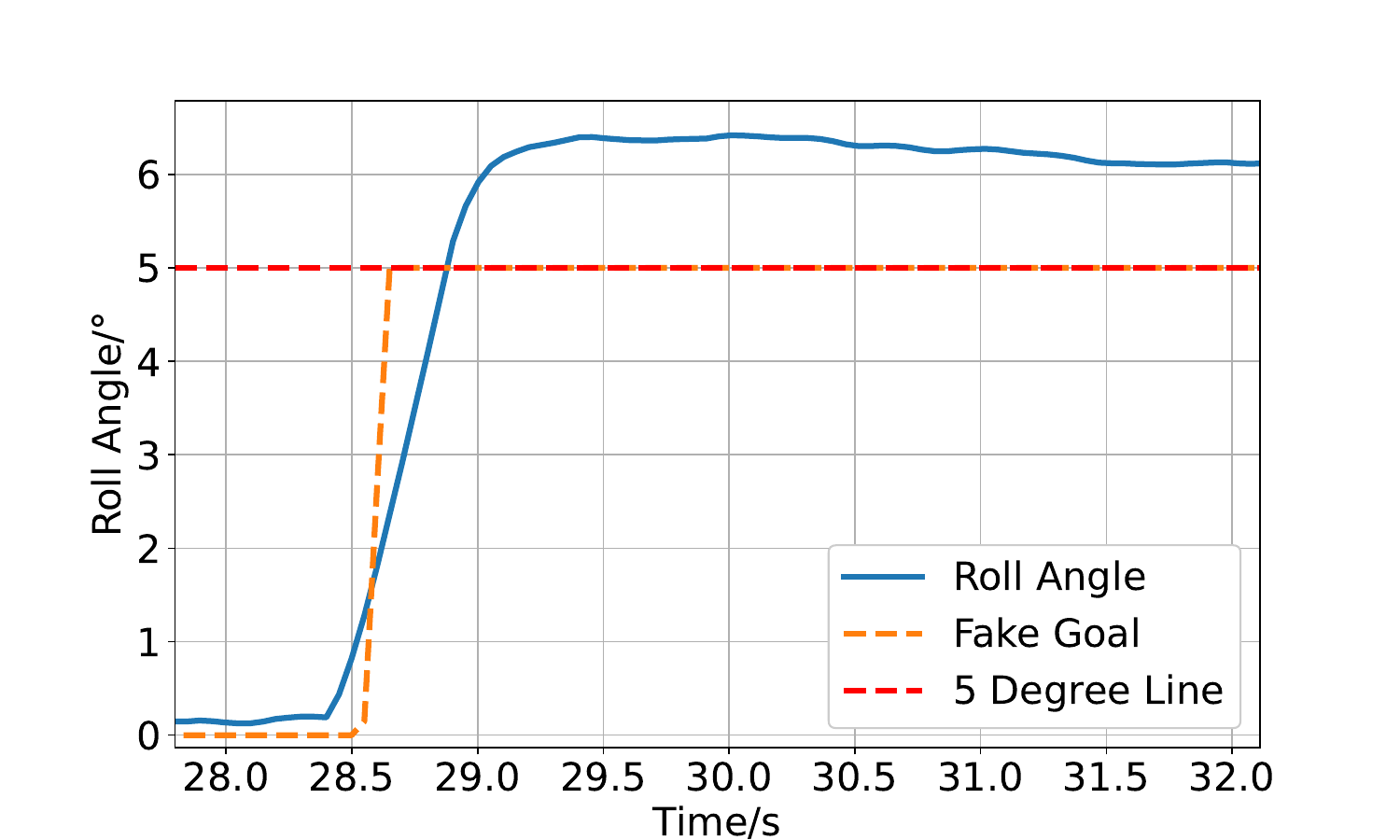}
       \caption{Without adviser: Agent struggles to reach target but fails.}
       \label{fig:attitude_control_without_adviser}
   \end{subfigure}
   \hfill
   \begin{subfigure}[b]{0.35\textwidth}
       \includegraphics[width=\textwidth]{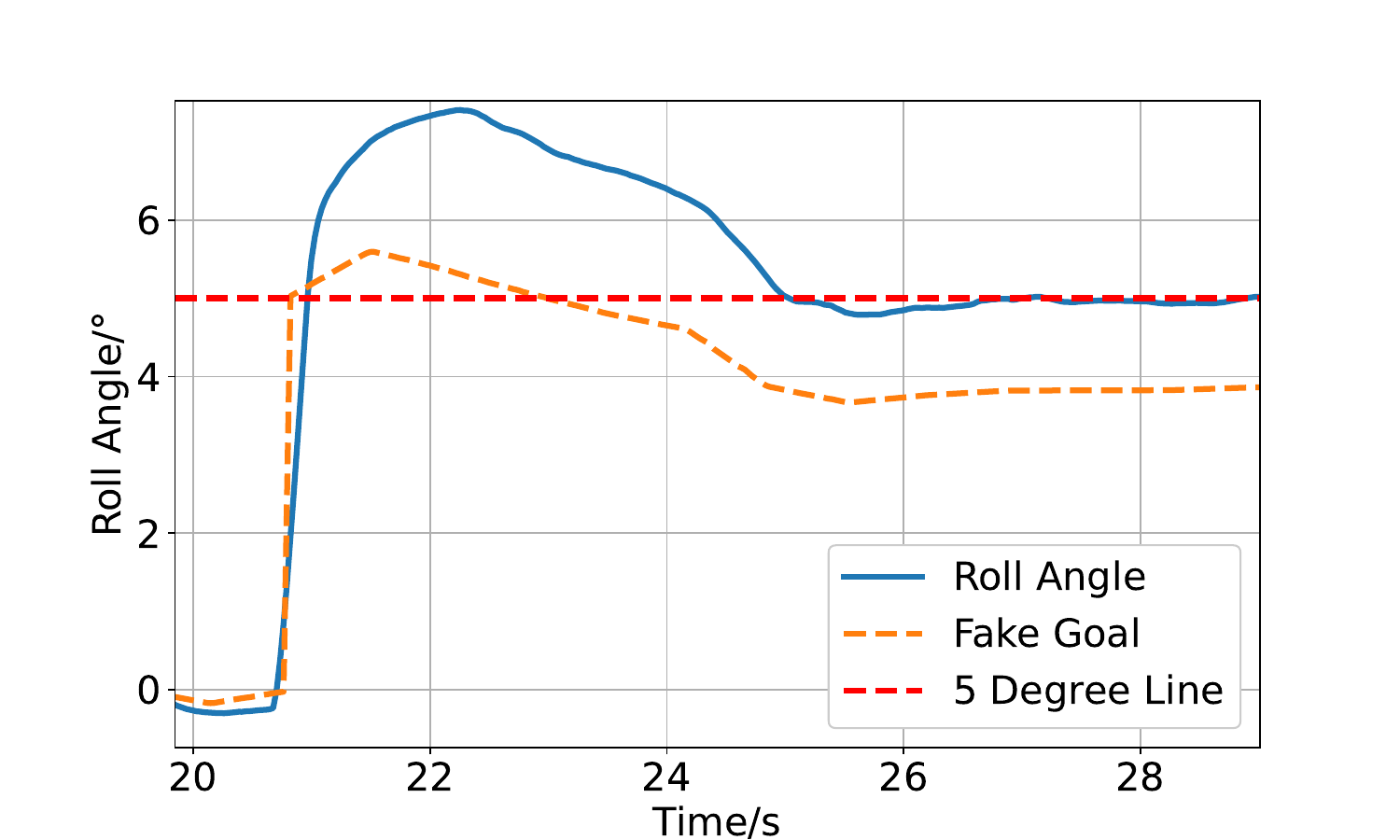}
       \caption{With adviser: Agent achieves target precisely.}
       \label{fig:attitude_control_with_adviser}
   \end{subfigure}
   \caption{Comparison of roll angle control with and without an adviser, highlighting improved precision and efficiency.}
   \label{fig:attitude_control_comparison}
\end{figure}

Further insight into the performance of the attitude control system is provided by Figure~\ref{fig:attitude_control_comparison}. The experiment involved targeting a specific roll angle of 5° from an initial roll angle of 0°, while keeping pitch and yaw angles constant at 0°. The inclusion of an adviser during the training process of the SAC algorithm significantly enhances the precision of the control system, as evidenced by the smoother transition and more accurate attainment of the desired roll angle.

In summary, the experimental deployment on a real-world quadcopter has shown that Adviser-Actor-Critic framework can markedly improve the system's responsiveness and overall performance. This advancement underscores the potential for developing more sophisticated and dependable UAV (Unmanned Aerial Vehicle) control systems for a variety of applications.

\section{Discussion and Conclusion}

The Adviser-Actor-Critic (AAC) framework combines classical control theory with reinforcement learning to enhance control precision in complex systems. By using an adviser based on PID principles to guide the actor, AAC improves precision, reliability, and robustness in high-precision tasks. This approach effectively reduces steady-state errors and enhances exploration and training efficiency, showing significant improvements in environments like robotics arm and quadcopter control. AAC's effectiveness extends from simulations to real-world applications, making it valuable for advanced control systems in robotics and complex mechanical systems.

\textbf{Future Works.} Looking ahead, the AAC framework holds substantial potential for further development. The design of the adviser component is particularly ripe for exploration. Beyond methods such as Model Predictive Control (MPC) and H-infinity ($H_\infty$) control, which can reduce both transient and steady-state errors, numerous other control strategies could be integrated into the adviser. These might include adaptive control, optimal control, or even hybrid approaches that combine multiple methodologies. Exploring these options could significantly broaden AAC's capabilities, enhancing its performance in a wide range of dynamic and uncertain environments. The ongoing refinement of the adviser promises to unlock new possibilities for achieving precise and reliable control in advanced applications.

\textbf{Impact Statements.} This paper presents work whose goal is to advance the field of Machine Learning. There are many potential societal consequences of our work, none which we feel must be specifically highlighted here.


\bibliography{references}
\bibliographystyle{icml2025}

\newpage
\appendix
\onecolumn
\section{Conditions for Setting Fake Goals to Reduce Steady-State Error}
\label{appendix:decoupled}

In the context of control systems, particularly in multi-input multi-output (MIMO) systems, reducing steady-state error is a critical objective. Steady-state error is minimized when the actor approximates an optimal controller; however, zero steady-state error can still be achieved by setting suitable fake goals, even if the actor does not fully serve this role. The conditions under which setting fake goals can reduce steady-state error are discussed below.

The $k$-th order derivative of the $i$-th component of the error vector $\boldsymbol{e}$ can be expressed as:
\begin{equation}
e_i^{(k)} = -\sum_{j=1}^{n} \sum_{l=0}^{k-1} a_{ijl} e_j^{(l)} + \sum_{j=1}^{n} b_{ij} u_j + d_i
\end{equation}

where:
\begin{itemize}
   \item $ e_i^{(k)} $ is the $ k $-th order derivative of the $ i $-th error component.
   \item $ a_{ijl} $ are the coefficients representing the influence of the $ l $-th derivative of the $ j $-th error component on the $ k $-th derivative of the $ i $-th error component.
   \item $ u_j = \varepsilon_j + e_j $ is the $ j $-th component of the input vector.
   \item $ b_{ij} $ are the coefficients representing the influence of the $ j $-th input vector on the $ k $-th derivative of the $ i $-th error component.
   \item $ d_i $ is the disturbance affecting the $ k $-th derivative of the $ i $-th error component.
\end{itemize}

Taking the Laplace transform with zero initial conditions, we obtain:

\begin{equation}
s^k E_i(s) = -\sum_{j=1}^{n} \sum_{l=0}^{k-1} a_{ijl} s^l E_j(s) + \sum_{j=1}^{n} b_{ij} U_j(s) + D_i(s)
\end{equation}

Here, $E_i(s)$ represents the Laplace transform of the $i$-th error signal, $U_j(s)$ represents the Laplace transform of the $j$-th input vector, and $D_i(s)$ represents the Laplace transform of the $i$-th disturbance.

This equation can be expressed in matrix form as:

\begin{equation}
\boldsymbol{A}(s) \boldsymbol{E}(s) = \boldsymbol{B} \boldsymbol{U}(s) + \boldsymbol{D}(s)
\end{equation}

where:

\[
\boldsymbol{A}(s) = \begin{bmatrix}
s^k + \sum_{l=0}^{k-1} a_{11l} s^l  & \cdots &  \sum_{l=0}^{k-1} a_{1nl} s^l \\
\vdots  & \ddots & \vdots \\
 \sum_{l=0}^{k-1} a_{n1l} s^l &   \cdots & s^k + \sum_{l=0}^{k-1} a_{nnl} s^l
\end{bmatrix}
\]

$\boldsymbol{A}(s)$ is the system matrix, where each element $a_{ijl}$ represents the coefficients of the error dynamics.

\[
\boldsymbol{B} = \begin{bmatrix}
b_{11}  & \cdots & b_{1n} \\
\vdots  & \ddots & \vdots \\
b_{n1}  & \cdots & b_{nn}
\end{bmatrix}
\]

$\boldsymbol{B}$ is the input matrix, where each element $b_{ij}$ represents the influence of the $j$-th input vector on the $i$-th error.

Solving for $\boldsymbol{E}(s)$:

\begin{equation}
\boldsymbol{E}(s) = \boldsymbol{A}(s)^{-1} [\boldsymbol{B} \boldsymbol{U}(s) + \boldsymbol{D}(s)]
\end{equation}

This equation provides the Laplace transform of the error signals $\boldsymbol{E}(s)$ in terms of the input vector $\boldsymbol{U}(s)$ and disturbances $\boldsymbol{D}(s)$, given the system matrices $\boldsymbol{A}(s)$ and $\boldsymbol{B}$.

When a system incorporates a qualified actor, it is expected to exhibit two primary properties: stability and command tracking. These properties are essential for ensuring that advisers work effectively.

\textbf{Stability} ensures that, with zero inputs and constant disturbances, the error within the system stabilizes to a constant value after a sufficiently long period, preventing unbounded growth in the system's response over time. Mathematically, stability is represented by the requirement that all poles of the system's transfer function $\boldsymbol{A}(s)^{-1}$ have negative real parts, ensuring that the system's transient responses decay over time, leading to a stable equilibrium:
\begin{equation}
   \text{Re}(\lambda) < 0 \quad \forall \lambda \in \text{poles of } \boldsymbol{A}(s)^{-1}
   \label{eq:qualified_condition1}
\end{equation}

\textbf{Command tracking} ensures that perturbation in the input vector result in corresponding changes in the error, similar to the perturbation. Mathematically, this property is represented by the condition that the change in the error vector $\boldsymbol{e}$ is similar to input perturbation. When input perturbation equals $-\boldsymbol{e}_{ss,\text{prev}}$, we have:
\begin{equation}
   \begin{aligned}
   \boldsymbol{e}_{ss,\text{now}} &= \lim_{s \to 0} s\boldsymbol{A}(s)^{-1} \left[\boldsymbol{B} \left(\boldsymbol{U}(s) - \frac{\boldsymbol{e}_{ss,\text{prev}}}{s}\right) + \boldsymbol{D}(s)\right] \\
   &= \lim_{s \to 0} s\boldsymbol{A}(s)^{-1} \left[\boldsymbol{B} \boldsymbol{U}(s) + \boldsymbol{D}(s)\right] - \lim_{s \to 0} \boldsymbol{A}(s)^{-1} \boldsymbol{B} \boldsymbol{e}_{ss,\text{prev}} \\
   &= \left(\boldsymbol{I} - \boldsymbol{B}\right)\boldsymbol{e}_{ss,\text{prev}}
   \end{aligned}
\end{equation}
where $\boldsymbol{e}_{ss,\text{prev}}$ denotes the original steady-state error, and $\boldsymbol{e}_{ss,\text{now}}$ denotes the new steady-state error.

Taking the 2-norm (or any other compatible matrix norm) on both sides, we get:
\begin{equation}
\|\boldsymbol{e}_{ss,\text{now}}\|_2 = \|(\boldsymbol{I} - \boldsymbol{B}) \boldsymbol{e}_{ss,\text{prev}}\|_2
\end{equation}

Using the sub-multiplicative property of matrix norms, we have:
\begin{equation}
\|\boldsymbol{e}_{ss,\text{now}}\|_2 \leq \|\boldsymbol{I} - \boldsymbol{B}\|_2 \cdot \|\boldsymbol{e}_{ss,\text{prev}}\|_2
\end{equation}

For any matrix norm that is compatible with the vector norm, the spectral radius $\rho(\boldsymbol{\boldsymbol{I} - \boldsymbol{B}})$ satisfies:
\begin{equation}
\rho(\boldsymbol{\boldsymbol{I} - \boldsymbol{B}}) \geq \|\boldsymbol{\boldsymbol{I} - \boldsymbol{B}}\|_2
\end{equation}
Given that $\rho(\boldsymbol{I} - \boldsymbol{B}) < 1$, there exists a compatible matrix norm such that:
\begin{equation}
\|\boldsymbol{e}_{ss,new}\|_2 \leq \|\boldsymbol{I} - \boldsymbol{B}\|_2 \cdot \|\boldsymbol{e}_{ss,\text{prev}}\|_2 \leq \rho(\boldsymbol{I} - \boldsymbol{B}) \cdot \|\boldsymbol{e}_{ss,\text{prev}}\|_2 < \|\boldsymbol{e}_{ss,\text{prev}}\|_2
\label{eq:qualified_condition2}
\end{equation}
We have shown that if $\rho(\boldsymbol{I} - \boldsymbol{B}) < 1$, then the 2-norm of the new steady-state error vector is strictly less than the 2-norm of the previous steady-state error vector. This implies that the error decreases with each iteration, indicating that modifying the inputs based on the original steady-state error vector $\boldsymbol{e}_{ss,\text{prev}}$ results in a reduced new steady-state error vector $\boldsymbol{e}_{ss,\text{now}}$:

In summary, a system incorporating an actor should exhibit both stability and command tracking. The conditions presented in Equations.~(\ref{eq:qualified_condition1}) and (\ref{eq:qualified_condition2}) facilitate the quantitative analysis of these properties and aid in designing systems that meet the required performance criteria. Special conditions apply when the Actor is an optimal controller.  Mathematically, this can be expressed as:

\begin{equation}
   Re(\lambda(\boldsymbol{A}(s) - \boldsymbol{I})) \ll 0, \quad \rho(\boldsymbol{I} - \boldsymbol{B}) \approx 0
   \label{eq:fully_decoupled}
\end{equation}

\section{Experiment Setting and Algorithm Configuration}
\label{appendix:experiment}
The following tables provide a detailed summary of the configurations for our three experimental environments, namely the mass-spring-damper System, the robotics arm, and the quadcopter. Each table outlines key aspects of the environment, including the description, action space, goal space, observation space, rewards, starting state, and episode termination conditions. These environments are designed to test and validate the performance of our control algorithms under different dynamics and constraints. Additionally, information on experiment parameters is included in following tables.

To further elaborate on the experimental setup of the quadcopter environment, we provide a detailed description of its configuration and operational framework. The primary drone used for data collection has a takeoff weight of 1.40 kg, with a quadrotor configuration that is center-symmetric and an axle distance of 35.0 cm. The rotor radius is 12.0 cm, and the rotors are driven by PWM motors. The low-level flight control system utilizes a Pixhawk4 (PX4) to generate PWM signals for hardware execution. The high-level system is based on the Robot Operating System (ROS), which provides interfaces for external information and commands. The PX4 flight control system natively relies on IMU and global positioning data (typically GPS). For indoor laboratory flights, a motion capture system is used instead of GPS to deliver more reliable position and attitude information.

\begin{table}[h]
   \caption{Environment configuration summary for mass-spring-damper.}
   \label{tab:env_config}
   \vskip 0.15in
   \begin{center}
   \begin{small}
   \begin{tabular}{ll}
   \toprule
   \textbf{Aspect} & \textbf{Explanation} \\ 
   \midrule
   \textbf{Description} & A 2D mass-spring-damper system where a controller applies force to maintain a desired position. \\ 
   \textbf{Action Space} & Forces applied in the $x$ and $y$ directions: $\mathbf{a} \in [-1.0, 1.0]^2$. \\ 
   \textbf{Goal Space} & Target position and velocity in 2D space: $\mathbf{g} \in \mathbb{R}^4$ (position in $[-2.4, 2.4]^2$ and velocity $[0.0, 0.0]^2$). \\ 
   \textbf{Observation Space} & Current position and velocity of the robot: $\mathbf{o} \in \mathbb{R}^4$ (position in $[-2.4, 2.4]^2$ and velocity in $[-0.1, 0.1]^2$). \\ 
   \textbf{Rewards} & 
   $
   r = -1.0 \cdot \|\mathbf{o}_{\text{position}} - \mathbf{g}_{\text{position}}\|^2 - 0.5 \cdot \|\mathbf{o}_{\text{velocity}} - \mathbf{g}_{\text{velocity}}\|^2 - 0.1 \cdot \|\mathbf{a}\|
   $
   \\ 
   \textbf{Starting State} & $\mathbf{s}_0 \sim \text{Uniform}([-2.4, 2.4]^2 \times [-0.1, 0.1]^2)$ \\ 
   \textbf{Episode Termination} & $|x| > 4.8$ $||$ $|y| > 4.8$, or  $\sqrt{x^2+y^2} < 0.01$ \&\& $\sqrt{v_x^2+v_y^2} < 0.01$. \\ 
   \textbf{Episode Truncation} & Number of steps $\geq 1000$. \\ 
   \bottomrule
   \end{tabular}
   \end{small}
   \end{center}
   \vskip -0.1in
\end{table}

\begin{table}[h]
   \caption{Environment configuration summary for robotics arm.}
   \label{tab:arm_env_config}
   \vskip 0.15in
   \begin{center}
   \begin{small}
   \begin{tabular}{ll}
   \toprule
   \textbf{Aspect} & \textbf{Explanation} \\ 
   \midrule
   \textbf{Description} & A 3-DoF robotic arm where a controller applies torques to move the end-effector to a desired position. \\ 
   \textbf{Action Space} & Torques applied to each joint: $\mathbf{a} \in [-1.0, 1.0]^3$. \\ 
   \textbf{Goal Space} & Target end-effector position in 3D space: $\mathbf{g} \in \mathbb{R}^3$ (position in $[0.5, 1.0] \times [0.5, 1.0] \times [0.0, 0.3]$). \\ 
   \textbf{Observation Space} & Current joint angles (cosine and sine) and angular velocities: $\mathbf{o} \in \mathbb{R}^9$. \\ 
   \textbf{Rewards} & 
   $
   r = -1.0 \cdot \|\mathbf{o}_{\text{end-effector}} - \mathbf{g}\| - 0.1 \cdot \|\mathbf{o}_{\text{velocity}}\| - 0.1 \cdot \|\mathbf{a}\|
   $
   \\ 
   \textbf{Starting State} & $\mathbf{s}_0 \sim \text{Uniform}([-\pi, \pi]^3 \times [-0.005, 0.005]^3)$ \\ 
   \textbf{Episode Termination} & Joint velocities exceed or $\|g_o - g_a\| < 0.1$. \\ 
   \textbf{Episode Truncation} & Number of steps $\geq 1000$. \\ 
   \bottomrule
   \end{tabular}
   \end{small}
   \end{center}
   \vskip -0.1in
\end{table}
\begin{table}[h]
   \caption{Environment configuration summary for quadcopter.}
   \label{tab:quadcopter_env_config}
   \vskip 0.15in
   \begin{center}
   \begin{small}
   \begin{tabular}{ll}
   \toprule
   \textbf{Aspect} & \textbf{Explanation} \\ 
   \midrule
   \textbf{Description} & A quadcopter where a controller adjusts motor thrusts to achieve a target velocity. \\ 
   \textbf{Action Space} & Vertical velocity setpoint and attitude setpoints (roll, pitch, yaw): $\mathbf{a} \in [-2.0, 2.0] \times [-0.2, 0.2]^3$. \\ 
   \textbf{Goal Space} & Target velocity in 3D space: $\mathbf{g} \in \mathbb{R}^3$ (velocity in $[-1.0, 1.0]^3$). \\ 
   \textbf{Observation Space} & Current velocity, acceleration, angular rates, and attitude angles: $\mathbf{o} \in \mathbb{R}^{12}$. \\ 
   \textbf{Rewards} & 
   $
   r = -\|\mathbf{o}_{\text{velocity}} - \mathbf{g}\|
   $
   \\ 
   \textbf{Starting State} & Initial velocity and attitude randomly sampled (velocity in $[-1.0, 1.0]^3$, attitude near zero). \\ 
   \textbf{Episode Termination} & Joint velocities exceed or $\|g_o - g_a\| < 0.05$. \\ 
   \textbf{Episode Truncation} & Number of steps $\geq 1000$. \\ 
   \bottomrule
   \end{tabular}
   \end{small}
   \end{center}
   \vskip -0.1in
\end{table}

\begin{table}[h]
   \caption{Details of the experiment parameters for robotic arm}
   \vskip 0.15in
   \begin{center}
   \begin{tabular}{clcccccc}
   \toprule
   \multirow{2}{*}{Strategy}&\multirow{2}{*}{Adviser} & \multicolumn{3}{c}{adviser train params} & \multicolumn{3}{c}{adviser eval params}\\
   \cmidrule(lr){3-5}\cmidrule(lr){6-8}
   && \raisebox{-1pt}{$K_p$} & \raisebox{-1pt}{$K_i$} & \raisebox{-1pt}{$K_d$} & \raisebox{-1pt}{$K_p$} & \raisebox{-1pt}{$K_i$} & \raisebox{-1pt}{$K_d$ } \\ 
   \midrule
   \multirow{4}{*}{SAC} & No adviser & 1.0 & 0.0 & 0.0  & 1.0 & 0.0 & 0.0 \\
   &Evaluate adviser & 1.0 & 0.0 & 0.0 & 1.3 & 0.1 & 0.1 \\
   &Train adviser & 1.3 & 0.01 & 0.01 & 1.0 & 0.0 & 0.0 \\
   &\makecell[l]{Train + evaluate adviser} & 1.3 & 0.01 & 0.01 & 1.3 & 0.1 & 0.1 \\
   \midrule
   \multirow{4}{*}{\makecell{SAC\\+\\HER}} & No adviser & 1.0 & 0.0 & 0.0 & 1.0 & 0.0 & 0.0 \\
   &Evaluate adviser & 1.0 & 0.0 & 0.0 & 1.3 & 0.1 & 0.1 \\
   &Train adviser & 1.3 & 0.01 & 0.01 & 1.0 & 0.0 & 0.0 \\
   &\makecell[l]{Train + evaluate adviser} & 1.3 & 0.01 & 0.01 & 1.3 & 0.1 & 0.1 \\
   \bottomrule
   \end{tabular}
   \end{center}
\end{table}

\begin{table}[h]
   \caption{Common hyperparameters for all experiments}
   \vskip 0.15in
   \begin{center}
   \begin{tabular}{lc}
   \toprule
   Hyperparameter & Value \\
   \midrule
   Number of epochs & 51 \\
   Initial value for alpha & 0.2 \\
   Discount factor for future rewards & 0.995 \\
   Learning rate for Q networks & 5e-4 \\
   Learning rate for actor network & 3e-4 \\
   Learning rate for alpha & 3e-4 \\
   Tau for soft updates & 0.005 \\
   Dimension of hidden layers & 128 \\
   Minimum replay buffer size before learning starts & 1000 \\
   Maximum replay buffer size & 1e6 \\
   Batch size for training & 64 \\
   Number of hidden layers & 3 \\
   Activation function for hidden layers & SELU \\
   \bottomrule
   \end{tabular}
   \end{center}
\end{table}

\end{document}